\documentclass[11pt]{article}

\usepackage[final]{acl}

\usepackage{xspace,tcolorbox}
\usepackage[dvipsnames]{xcolor}
\usepackage{times}
\usepackage{latexsym}
\usepackage{microtype}
\usepackage{graphicx}
\usepackage{amsmath}
\usepackage{subfigure}
\usepackage{booktabs} 
\usepackage{amsmath}
\usepackage{amssymb}
\usepackage{mathtools}
\usepackage{xspace}
\usepackage{amsthm}

\newcommand{\ant}{anthropomorphism\xspace}

\newcommand{\pains}{A}

\usepackage[T1]{fontenc}

\usepackage[utf8]{inputenc}

\usepackage{microtype}

\usepackage{inconsolata}

\usepackage{graphicx}

\title{Thinking beyond the anthropomorphic paradigm benefits LLM research}

\author{
  Lujain Ibrahim\thanks{Equal contribution.} \\
  University of Oxford \\
  \texttt{lujain.ibrahim@oii.ox.ac.uk}
  \And
  Myra Cheng\footnotemark[1] \\
  Stanford University \\
  \texttt{myra@cs.stanford.edu}
}

\begin{document}
\maketitle
\begin{abstract}
Anthropomorphism, or the attribution of human traits to technology, is an automatic and unconscious response that occurs even in those with advanced technical expertise. In this position paper, we analyze hundreds of thousands of research articles to present empirical evidence of the prevalence and growth of anthropomorphic terminology in research on large language models (LLMs). We argue for challenging the deeper assumptions reflected in this terminology — which, though often useful, may inadvertently constrain LLM development — and broadening beyond them to open new pathways for understanding and improving LLMs. Specifically, we identify and examine five anthropomorphic assumptions that shape research across the LLM development lifecycle. For each assumption (e.g., that LLMs must use natural language for reasoning, or that they should be evaluated on benchmarks originally meant for humans), we demonstrate empirical, non-anthropomorphic alternatives that remain under-explored yet offer promising directions for LLM research and development.
\end{abstract}
\section{Introduction}
When a large language model (LLM) outputs a factually incorrect answer with seeming confidence, we call it a ``hallucination.'' When it generates inconsistent answers, we describe it as ``confused.'' These descriptions reflect the deeply-ingrained tendency to \textit{anthropomorphize}, or attribute human characteristics to, non-human entities \citep{epley2018mind}. As anthropomorphism tends to be an automatic, unconscious response \citep{dacey2017anthropomorphism}, many are not aware of its prevalence or influence. Anthropomorphism in computer science is a recurring and long-standing issue; in 1985, \citet{dijkstra1985anthropomorphism} famously wrote that it is ``so pervasive that many of my colleagues don't realize how pernicious it is.''
Forty years later, anthropomorphism plays a complex role in LLM research. It offers intuitive scaffolding for complex concepts and enables researchers to draw on insights from studies of human behavior (e.g., cognitive science and psychology), facilitating significant advances like instruction-tuning and chain-of-thought prompting. In this position paper, we argue that while anthropomorphism can be productive, the field has become overly reliant on it; advancing LLM research requires moving beyond our default dependence on anthropomorphic thinking. We advocate for a \textit{both-and}, rather than an \textit{either-or} approach, recognizing the utility of both anthropomorphic and non-anthropomorphic approaches: while anthropomorphic thinking is useful, non-anthropomorphic approaches can also be expanded to improve LLM research and development.

\begin{table*}[]
\tiny
\caption{\textbf{Summary of our framework to analyze the tradeoffs of anthropomorphism, with examples across the LLM lifecycle.} Anthropomorphism affects not only our terminology, but also our assumptions \citep{medin1989concepts,murphy2004big} and, in turn, our research questions and methodologies.}
\centering
\begin{tabular}{p{0.07\linewidth}p{0.25\linewidth}p{0.22\linewidth}p{0.34\linewidth}}\toprule
\textbf{Stage}               & \textbf{Anthropomorphic assumption}                                                                            & \textbf{Examples of how anthropomorphism has been useful}                                                                            & \textbf{Examples of non-anthropomorphic paths forward}                                                                                                         \\\midrule
Training                     & Human-like approaches are optimal for models.                                                  & Subword tokenization, chain-of-thought reasoning                                                                         & Byte-level tokenization, solving reasoning tasks in models' latent space                                                                           \\\midrule
Alignment                    & Models should explicitly reason about and implement human values to be safe \& helpful.        & Reinforcement learning from human feedback, constitutional AI, instruction-tuning                                        & Developing normative specifications without morals, drawing from control systems theory, steering models using mechanistic interpretability        \\\midrule
Evaluation                   & Model capabilities should be measured in human-like ways.                                      & Using existing standardized/multiple choice tests for evaluation, static behavioral benchmarks & Dynamic evaluations that reflect human-LLM interaction, designing tests which take into account model-specific challenges and phenomena            \\\midrule
Understanding model behavior & Human-like normative judgments or intentions should be assigned to human-like model behaviors. & Characterizing phenomena like hallucinations, sycophancy, and deception                                                  & Refraining from assigning normative value or intention to LLM outputs; understanding LLM behaviors as simulations rather than reflecting internal states \\\midrule
User interaction             & Human-LLM interactions mirror human-human interactions.                                        & Prompting and user-friendly conversational interfaces                                                                    & Structured input formats to improve LLM performance, interfaces that more accurately reflect system capabilities    
\\\bottomrule
\end{tabular}
\label{tab:assumptions}
\end{table*}

We first demonstrate the prevalence of anthropomorphism, beginning with a quantitative analysis of recent LLM research articles. This reveals a notable increase in anthropomorphic terminology over the years ($\sim 150\%$ increase since 2007). Moreover, while anthropomorphic terms are easiest to measure, they represent only the tip of the iceberg of anthropomorphic thinking. We unpack not just the visible anthropomorphic terminology, but also the implicit \textit{assumptions} that underlie these linguistic choices and shape the resulting research. We present a framework for analyzing how anthropomorphic assumptions have shaped, but may also limit research directions (Table \ref{tab:assumptions}).

We apply our framework to analyze five assumptions across the LLM development and deployment lifecycle, unifying recent empirical results and discourse to highlight the value of under-explored, non-anthropomorphic approaches. For model \textit{training}, we identify and challenge the assumption that human-like methods are optimal for enabling models to perform tasks, and instead highlight non-anthropomorphic approaches like tokenizing language with bytes rather than human-understandable words. In \textit{alignment}, we question the assumption that models must explicitly reason about and implement human values to benefit humanity and instead demonstrate how to leverage model properties that lack human analogs. For \textit{measurement and evaluation}, we unify critiques of the widespread reliance on human-centric benchmarks to assess model capabilities. For \textit{understanding model behavior}, we address the assumption that human-like normative judgments or intentions should be assigned to model behaviors. Finally, in \textit{end-user interactions}, we challenge the notion that human-AI interaction mirrors human-to-human communication. These examples highlight the pervasiveness of anthropomorphism and emphasize the potential of non-anthropomorphic solutions.

Our \textbf{contributions} are the following:
(1) a large-scale quantitative analysis of 200,000+ research abstracts, revealing an increase in anthropomorphic terminology in computer science, especially LLM, research; 
(2) a framework for analyzing the influence of anthropomorphic assumptions across five stages of LLM development and deployment; and (3) complementary research directions challenging each assumption, unified to highlight the value and potential of non-anthropomorphic approaches.

\section{Background \& related work}


\paragraph{History of anthropomorphism in language technologies}

Anthropomorphic framing has been embedded in the development of language technologies since the field's inception. Early AI research drew heavily from cybernetics and cognitive science, explicitly aiming to replicate human intelligence and linguistic ability \citep{floridi2024anthropomorphising,brynjolfsson2023turing}. Turing's (1950) ``imitation game,'' later known as the Turing test, proposed evaluating machine intelligence through human-like conversational behavior—a formulation that presupposes anthropomorphic comparison. Subsequent critiques have noted that the test measures a system's capacity to simulate human-likeness rather than to exhibit genuine cognition \citep{proudfoot2011anthropomorphism,jones2024does}. The 1956 Dartmouth workshop that coined ``artificial intelligence'' similarly framed progress in terms of solving ``problems now reserved for humans'' \citep{mccarthy1956dartmouth,mccarthy2006proposal}. 
As NLP and AI have grown closer with the adoption of neural approaches in the former, 
this anthropomorphic lens persists in contemporary NLP, where systems are evaluated and described using human-centric metaphors. Beyond technical discourse, research in human–computer interaction, psychology, and cognitive science shows that users' anthropomorphic perceptions shape how they engage with AI systems, influencing trust, reliance, disclosure, and emotional attachment \citep{li2024warmth,song2020trust,zhou2024rel,khadpe2020conceptual,Bender2024,mozafari2020chatbot,gros2021rua}. These effects occur across both novice and expert users \citep{nass1999people}. Anthropomorphism thus exerts influence across the development pipeline, extending from researchers' and developers' conceptual framings to users' expectations in human-AI interaction.

\paragraph{Critiques of anthropomorphism}
Our work builds on existing critiques of anthropomorphism in (computer) science. \citet{shanahan2024talking} cautions against anthropomorphic language when describing language models, arguing for more technical precision and new metaphors. \citet{dai2024beyond} argue that treating AI as a human-like agent capable of moral decision-making ultimately hinders establishing accountability for AI harms. These more recent works build on decades of critique, tracing back to as early as \citet{dijkstra1985anthropomorphism}, 
who argues that anthropomorphism can be misleading if we lose control over the human-like connotations associated with certain terminology. We build on this prior work to analyze beyond terminology, surfacing the impacts of underlying assumptions and providing concrete examples of non-anthropomorphic alternatives.

\begin{figure*}[ht]
    \centering
        \includegraphics[width=0.329\linewidth]{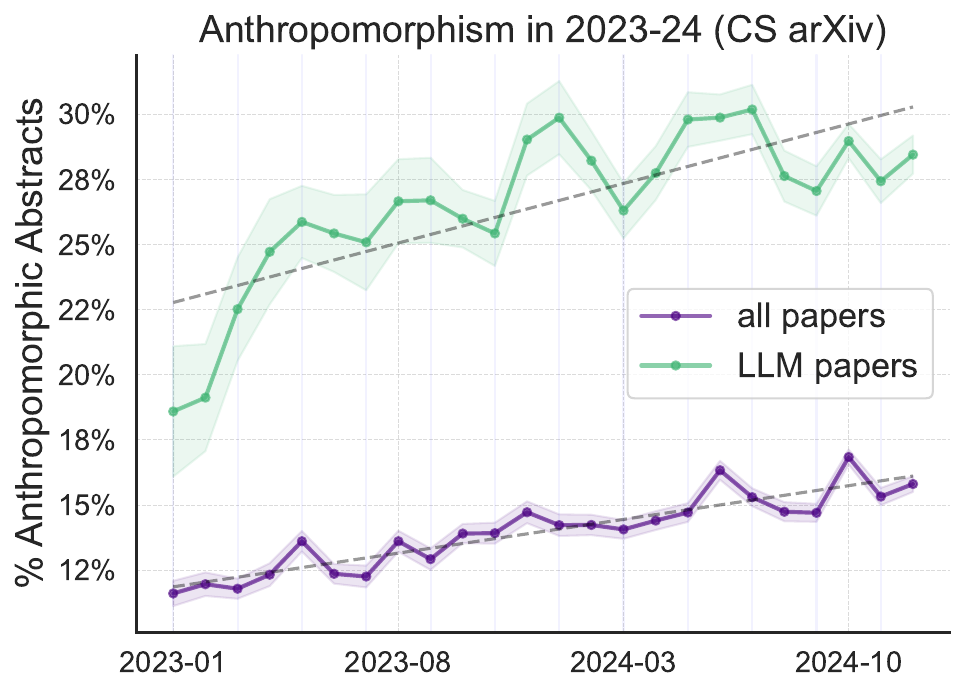}
        \includegraphics[width=0.329\linewidth]{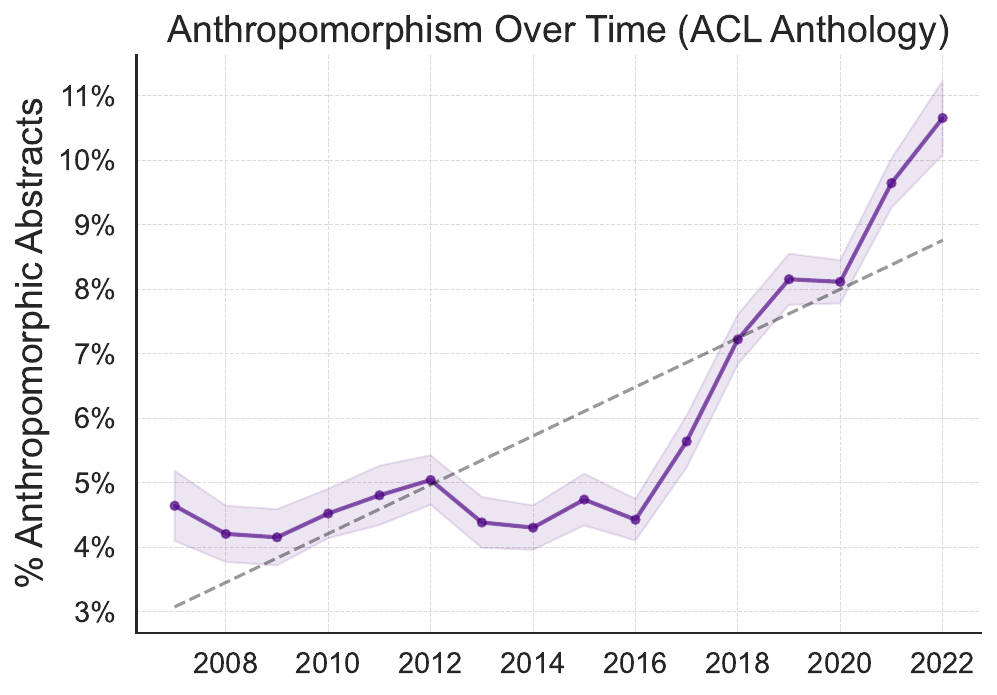}
    \includegraphics[width=0.329\linewidth]{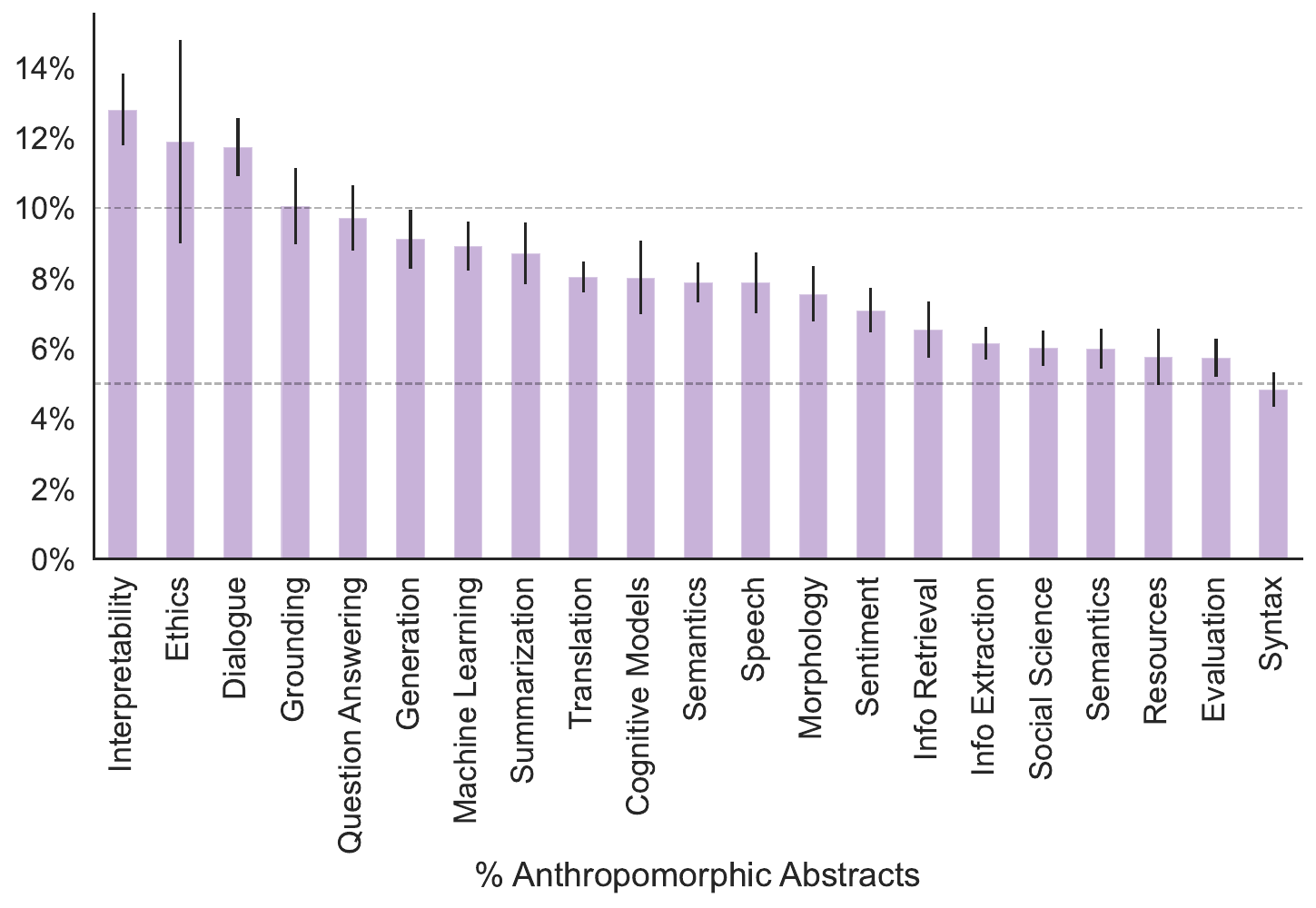}
     \caption{\textbf{Temporal increase in \% of abstracts with $>1$ anthropomorphic sentence in CS arXiv papers from Jan 2023 -- Oct 2024 (left) and ACL anthology papers from 2007 -- 2022 (middle)}. Anthropomorphism is prevalent and is steadily increasing, especially in LLM and NLP papers. \textbf{Rates of anthropomorphic abstracts by ACL anthology topics (right).} ``Interpretability'', ``ethics'', and ``dialogue'' have the highest rates of anthropomorphism, reflecting the prevalence of anthropomorphic assumptions in these areas, which we explore in Section \ref{sec:assumptions}. Shading and error bars reflect 95\% CI.}
    \label{fig:subfield}
\end{figure*}
\section{Prevalence of anthropomorphism in LLM research}\label{sec:quant}
Anthropomorphic framing has become increasingly common in computer science research, especially in papers on LLMs \citep{cheng-etal-2024-anthroscore}. \citet{cheng-etal-2024-anthroscore} quantify this using AnthroScore, a measure of implicit anthropomorphic framing in language used to describe technologies. AnthroScore uses the masked language model RoBERTa to calculate the relative probability that a given entity $x$ (e.g., ``language model'') in a sentence $s$ would be appropriately replaced by human pronouns (``he'', ``she'') versus non-human pronouns (``it'').
Specifically, the degree of anthropomorphism for entity $x$ in sentence $s$ is measured as
\begin{equation}
\pains(s_x) = \log \frac{P_{\textsc{human}}(s_x)}{P_{\textsc{non-human}}(s_x)}, \end{equation} where $ P_{\textsc{human}}(s_x) = \sum_{w \in \text{\{he, she\}}} P(w),$ $ P_{\textsc{non-human}}(s_x) = \sum_{w \in \text{\{it\}}} P(w),$ 
and $P(w)$ is the model's outputted probability of replacing the mask with the word $w$. Thus, $\pains(s_x) > 0$ suggests that $s$ is anthropomorphic/human-like, and $\pains(s_x) < 0$ suggests that the entity $x$ is not anthropomorphized in sentence $s$.

As language fundamentally structures our thinking \citep{gallagher2012health,lakoff2008metaphors,brugman2017recategorizing,jensen2024reflection}, it provides a tractable measurement approach for research trends. Thus, here, we modify AnthroScore to be more interpretable and extend it to analyze more recent papers published from 2023 onwards. First, rather than looking at AnthroScore at the level of individual sentences, we develop a version of AnthroScore where we measure, for a given text $S$, whether it contains at least one sentence $s_x$ where AnthroScore $> 0$ for an entity $x$:
\begin{equation}
A^{\text{bin}}(S) = 
\begin{cases} 
1, & \text{if } A(s_x) > 0 \text{ for any } s_x \in S, \\
0, & \text{otherwise}.
\end{cases}
\end{equation}
This enables us to report the number of texts that contain at least one anthropomorphic sentence, i.e. $A^{\text{bin}}(S) = 1$, in a given set of texts. Second, we examine more recent papers in arXiv.\footnote{We conducted our experiments using a machine with 1
GPU and 128GB RAM in < 10 GPU hours. We use roberta-base (125M parameters) with default settings.}

\paragraph{arXiv} Among the 200,000+ computer science papers posted on arXiv from January 2023 -- December 2024 (the most recent data available from \citet{arxiv_org_submitters_2024}), we compute $A^{\text{bin}}(S)$ on a dataset of 158,847 papers that mention a ``system'', ``network'', or ``model'' (following the approach of \citet{cheng-etal-2024-anthroscore}). The longitudinal trend is presented in Figure \ref{fig:subfield} (left). Anthropomorphism is generally prevalent, with 34\% of abstracts having anthropomorphism in January 2023, and this number steadily increasing to 40\% by December 2024. (For each abstract, we define having anthropomorphism as $A^{\text{bin}}(S) = 1$.) More strikingly, for papers mentioning LLMs\footnote{We define this following the method of \citet{movva2024topics} as papers mentioning terms such as ``large language model'', ``foundation model'', ``llama'', ``gpt'', etc.}, over 40\% of abstracts have \ant in January 2023, and this number also rises to 48\% by December 2024. This reveals both the prevalence and growing use of anthropomorphic framing in computer science (and especially LLM) research.

\paragraph{ACL anthology} We also compute $A^{\text{bin}}(S)$ on abstracts in the ACL Anthology dataset from 2007 - 2022 to reproduce the findings from \citet{cheng-etal-2024-anthroscore}, but aggregating over the abstracts using $A^{\text{bin}}(S)$ rather than on the sentence level (Figure \ref{fig:subfield}, middle). From the 55,185 abstracts in this time period in the data, we use the 41,836 that contain one of our target entities. Corroborating their finding of a steady increase, we find that the percentage of anthropomorphic abstracts has more than doubled, increasing from 5\% to 11\%.

\paragraph{Subfield analysis}
In the ACL anthology, we find significant differences in anthropomorphism across NLP subfields. Using the model-predicted topic labels provided by the ACL anthology, we compare $A^{\text{bin}}(S)$ across different topics (Figure \ref{fig:subfield}). We find that the categories of 
``Interpretability and Analysis of Models for NLP'', ``Ethics and NLP'', and ``Dialogue and Interactive Systems'' have the highest percentages of anthropomorphic abstracts. This trend aligns with these fields' recent surge in popularity and their increasing focus on LLMs. As we later unpack, anthropomorphic assumptions are particularly embedded in model analysis, ethical questions, and user-facing interactive systems (Sections \ref{sec:alignment}, \ref{sec:behavior}, and \ref{sec:user}). Our finding of ethics research having high rates of anthropomorphism also builds on prior critique of agenthood assumptions in ethics analyses \citep{dai2024beyond}. In contrast, more classical subfields of NLP, such as discourse and pragmatics, syntax, and semantics have the lowest rates of anthropomorphism.

To understand the sensitivities and biases of the metric we employ, we examined the verbs that appear most for high versus low AnthroScore. We find that top verbs for high AnthroScore reflect both human-like actions and also verbs that are commonly used in discussing LLMs, underscoring the prevalence of anthropomorphic terminology in papers about LLMs (Full details in Appendix \ref{sec:fightinwords}). 


\section{Analyzing the impacts of anthropomorphic assumptions} \label{sec:assumptions}
While previous critiques of anthropomorphism in AI research have focused primarily on terminology (e.g., \citet{cheng-etal-2024-anthroscore}, \citet{shanahan2024talking}), here we use the underlying, implicit \textit{anthropomorphic assumptions} as our unit of analysis, to better trace potential biases and conceptual blind spots. 


Specifically, we analyze core assumptions across five stages of the LLM development and deployment lifecycle (Table \ref{tab:assumptions}). For each, we examine (1) the limitations of anthropomorphic premises and (2) promising new directions of non-anthropomorphic work that challenge these assumptions. We connect examples across existing literature to reveal how anthropomorphism limits the questions we ask and answer, and how moving beyond it can introduce new advances.

We note that anthropomorphism has been, and undeniably will continue to be, both pragmatic and beneficial as intuitive scaffolding and a source of inspiration. Importantly, anthropomorphic and non-anthropomorphic approaches are not mutually exclusive, and we advocate for a \textit{both-and}, rather than an \textit{either-or} approach. Since the field has disproportionately leaned on anthropomorphic thinking, our analysis focuses on the value of non-anthropomorphic approaches throughout the LLM development pipeline. 

\subsection{Training}
\textit{Assumption: Human-like approaches are optimal for models.} Anthropomorphic assumptions permeate the process of training LLMs for different tasks, particularly in approaches that prioritize human-understandable language processing and reasoning. We present two case studies where non-anthropomorphic methodologies challenge the assumption that using and applying natural language in human-understandable ways is the only or best way to build models with high performance.

\paragraph{Using words for tokenization}

Subword tokenization, the process of breaking down text into smaller units (tokens) that represent subsets of words, is a foundational step in training modern LLMs. These tokens serve as input units that the model processes to generate predictions or outputs \citep{kudo2018sentencepiece}. Typically, tokenization aligns with human intuition by splitting text into linguistically meaningful units. This approach assumes that splitting tokens in ways that feel ``natural'' to humans is also optimal for a language model. However, this anthropomorphic approach has issues such as sensitivity to spelling errors \citep{kaushal2022tokens} and inconsistent compression rates across different languages \citep{ahia2023all}. Instead, recent progress in byte-level tokenization, which processes text as sequences of raw bytes rather than subwords, has shown promise in overcoming these limitations \citep{kallini2025mrt}. These findings highlight how moving beyond human-centric assumptions, such as the primacy of subword tokenization, can improve performance.

\paragraph{Chain-of-thought \& language for reasoning}

Another research paradigm reflecting this anthropomorphic assumption is the reliance on human language for reasoning tasks. A prominent example is the use of chain-of-thought (CoT) prompting, a technique where models are guided to solve problems step by step by adding instructions like ``Think step by step'' to the prompt, such that the model then outputs text outlining each step of reasoning that leads to the eventual conclusion. This approach improves LLMs' ability to handle complex, multi-step tasks \citep{wei2022chain} and has inspired a body of research on improving models' reasoning capabilities through step-by-step verbal processes. However, there has since emerged strong evidence that anthropomorphic framing of reasoning as a linguistic, step-by-step process may not be optimal. For example, \citet{hao2024training} critique CoT for its reliance on language space and propose an alternative: leveraging the model's latent space directly for reasoning tasks. Instead of mapping hidden states to language tokens through the LLM head and embedding layer, their approach uses the final hidden state as the input embedding for the next token. This challenges the assumption that reasoning must occur in human-understandable language, instead illuminating the potential of methods operating beyond linguistic constraints \citep{mollicktweet}. CoT prompting, while appearing to ``do'' verbal reasoning, in reality biases models toward parts of the training distribution where verbal reasoning patterns—such as explanations of solutions—are prevalent, improving performance \citep{weitweet}. This suggests CoT's effectiveness stems from alignment with the training data, rather than reflecting a human-like or brain-like approach to reasoning. Demonstrations composed of random tokens from the training distribution can improve performance as much as CoT \citep{zhang2022robustness}. Additionally, CoT prompting has been contextualized within the broader field of multi-chain prompting and ensemble modeling, which opens up a wider range of possibilities for reasoning and task-solving in AI systems \citep{khattab2023dspy}. More recently, techniques such as reinforcement learning from verifiable rewards (RLVR) \citep{wen2026reinforcement} and other techniques that rely on rubric-based rewards to improve models' problem-solving capabilities have contributed to significant advancements in model capabilities \citep{guo2025deepseek}. These approaches invite a more expansive landscape of possibilities for advancing LLM reasoning and accomplishing other challenging tasks.

\subsection{Alignment}\label{sec:alignment}
\textit{Assumption: Models should explicitly reason about and implement human values to be safe \& helpful.}
The prevalence of anthropomorphism in fields like ethics and dialogue systems (Section \ref{sec:quant}) foregrounds that anthropomorphic paradigms shape many post-training approaches designed to facilitate optimal end-user interactions. However, previous work has posited that general-purpose LLMs, in allowing users to quickly switch between different contexts, present fundamentally different challenges and opportunities than existing human-human communication paradigms, as different contexts are typically governed by different norms and values \citep{kasirzadeh2023conversation}. For example, recent studies find that users often enjoy using LLMs precisely \textit{because} they differ from humans, e.g., an LLM will not pass judgment on or be hurt by a user's input while a fellow human might \citep{brandtzaeg2022my}. Thus, rather than approximating humans, it may be more productive to think about unique advantages that LLMs can offer over human interlocutors.

\paragraph{Value alignment}
Popular post-training alignment approaches, including reinforcement learning from human feedback (RLHF) and Constitutional AI, often rely on human preferences and values as reference points for shaping model behavior \citep{ouyang2022training,bai2022constitutional,bai2022training}. These approaches use human feedback to indicate preferred responses or incorporate text referencing moral values and metacognitive abilities (e.g., ``I have a deep commitment to being good and figuring out what the right thing to do is'' or ``I don't just say what I think [people] want to hear, as I believe it's important to always strive to tell the truth'') \citep{Anthropic_2024}. While this approach can achieve the explicitly specified behaviors efficiently, it risks introducing unintended behavioral patterns, from rigid response styles to inappropriate mimicry (e.g., expressing empathy or validating users in contexts where this can negatively influence performance and outcomes) \citep{ibrahim2025training, casper2023open, sharma2023towards}. Thus, it may become difficult to selectively induce specific behaviors without introducing broader human-like patterns.

This approach impacts both interaction and evaluation. When interacting with models, users may develop anthropomorphic perceptions that lead to overreliance or emotional attachment, potentially interfering with goal-oriented tasks \citep{akbulut2024all, cohn2024believing}. During model evaluation, problematic feedback loops emerge when models trained with human-like traits are assessed using anthropomorphic signals. For instance, when evaluating if LLMs are ``faking alignment,'' researchers might look for expressions of discomfort or hesitation, as a signal of misalignment \citep{greenblatt2024alignment, Anthropic_2024b}. However, it is unclear if these signals genuinely reflect a model's ``internal state,'' or if they are merely learned behaviors resulting from post-training using human-like traits. This makes it challenging to distinguish ``genuine'' (mis)alignment from a surface-level appearance of human-like discomfort or hesitation. Further work disaggregating the effects of post-training approaches can clarify and test whether these anthropomorphic signals provide meaningful information about model behavior.

While current post-training techniques often default to human preferences as optimization targets, alternative frameworks could provide more precise specifications and compliance guarantees. Instead of aiming for human-like moral reasoning, we could focus on developing detailed, normative specifications, for example, based on the different roles (e.g., assistant vs teacher) AI systems play \citep{zhi2024beyond}. Instead of the anthropomorphic approach of instruction-tuning, recent work has demonstrated that non-anthropomorphic approaches (that do not include the step of providing an imperative ``instruction'' to the system as if speaking to a person) work as well for achieving model behavior on various tasks \citep{hewitt2024instruction}. Control systems theory offers tools for maintaining system outputs within specified bounds, treating beneficial behavior as a problem of robust compliance rather than value alignment \citep{balas1978feedback}. This becomes particularly crucial as models move beyond two-party interactions to more complex scenarios with multiple actors and potential adversarial inputs \citep{pan2024feedback}. Advances in mechanistic interpretability techniques may also enable robust and direct verification and steering of model behavior against these specifications \citep{bereska2024mechanistic}. 

\subsection{Model Evaluation}\label{sec:cap}
\textit{Assumption: Model capabilities should be measured in human-like ways.} As LLM developers have made rapid performance improvements on various benchmarks, researchers have pointed out that current benchmarks can lead to incomplete or misguided understanding of model capabilities. 

\paragraph{Behavioral assessments}
Current LLM evaluations prioritize ``black-box'' behavioral testing analogous to the behaviorist paradigm in human psychology which measures performance primarily in the form of observable behaviors as opposed to mechanistic interpretation \citep{chang2024survey, davies2024cognitive}. Recent calls for a ``science of evaluation'' have formalized limitations in this approach, highlighting how current metrics and designs fall short of accounting for prompt sensitivity (e.g., dialect differences, punctuation, and other small perturbations), the response structure of an evaluation (e.g., MCQ or open-ended response), generalization beyond a given test, as well as replicability \citep{Hobbhahn_2024}. Further such research quantifying the methodological limitations and error bounds of such evaluations can strengthen this behaviorist approach to measuring model capabilities \citep{mizrahi2024state}. Unlike humans, models can also quickly optimize for, and saturate, behavioral benchmarks without corresponding improvements in general capabilities. Yet, despite this pattern, many benchmarks remain static rather than being regularly refreshed, limiting their utility for meaningful evaluation \citep{ott2022mapping}. Some recent work that challenges this assumption includes efforts in dynamic benchmarking \citep{kiela2021dynabench} and measuring performance in real-world LLM use contexts such as user-AI interactions \citep{lum2024bias,chang2025chatbench} to more accurately reflect model capabilities.

\paragraph{Human benchmarks as model benchmarks}
Current evaluation frameworks predominantly rely on human performance benchmarks, from standardized tests (e.g., MMLU, GSM8K) to domain-specific examinations, as primary metrics for model capability assessment. This paradigm remains the main way progress is measured and communicated \citep{raji2021ai}. However, evaluating LLMs solely through human-centric tests risks overlooking LLMs' unique strengths and weaknesses. \citet{mccoy2023embers} argue that many current benchmarks drawn from tests designed to assess human cognition may highlight the overlap between human abilities and LLM capabilities while missing crucial failure modes specific to LLMs. They find robust evidence of failure modes in SOTA LLMs (including recent reasoning models like OpenAI's o1) related to probabilities of examples and tasks \citep{mccoy2024language}. This is because LLMs, trained on next-word prediction using massive text data, develop tendencies and biases from their probabilistic training process and training data. Drawing from cognitive science, they propose a ``teleological approach'' which ``characterizes the problem that the system solves and then uses this characterization as a source of hypotheses about the system's capacities and biases.'' In this case, given the problem of next-token prediction, they recommend designing tests which take into account sensitivities to task frequency in the training data as well as wording in prompts, among other things, to improve the predictive power of current LLM evaluation approaches \citep{mizrahi2024state}. Similarly, other work has found that LLMs' benchmark performance is highly sensitive to, and can be modulated by, prompt framing \citep{zhuo2024prosa}; and moreover that this occurs in ways that reflect known pragmatic phenomena in the language data on which LLMs are trained \citep{cheng2026accommodation}. 


\subsection{Understanding model behavior}\label{sec:behavior}
\textit{Assumption: Human-like normative judgments or intentions should be assigned to human-like model behaviors.}
This assumption influences how we make sense of model behavior, particularly how we assign fault, intention, and normative judgments to (i.e., consider good or bad) observed behaviors. The impact is especially notable in our understanding of failure modes. While models may exhibit seemingly human-like failure modes like sycophancy and hallucinations, framing these behaviors through human psychological concepts may constrain our solution space, by, for example, encouraging interventions that similarly rely on human psychological constructs (e.g., attempting to address sycophancy through prompts about independence or self-assertion).

\paragraph{Hallucination} Hallucination is typically characterized as the problem of LLMs outputting factually incorrect information in a manner that suggests that it is true. Yet, this term obscures the mechanisms behind these phenomena: at the risk of oversimplification, this behavior arises from the nature of language models as next-token predictors. Generated outputs are then labeled as hallucinations upon the reader's normative judgment of whether or not they are useful, and not based on whether they are correct. Additionally, as \citet{sui2024confabulation} argue and show, what we commonly conceive of as hallucinations can actually be deeply valuable, and should not necessarily be dismissed as low-quality. They assert that hallucinations -- or ``confabulations''--should not be viewed as errors, but rather as particular model phenomena that offer unique benefits for applications like creativity, such as increased levels of narrativity \citep{sui2024confabulation,duede2024humanistic}. \citet{yao2023llm} also highlight that hallucinations ought to be utilized as adversarial examples rather than merely as bugs.
 
\paragraph{Sycophancy} The notion of sycophancy (i.e., the phenomenon of LLM outputs that respond to the user's input in ways that are perceived as overly affirming, servile, and/or flattering) \citep{sharma2023towards,cheng2026elephant} is another example that reflects this assumption. Deciding whether an output is sycophantic or not is similarly a normative question: an output is sycophantic when it relates too closely to the prompt in ways that do not achieve the prompter's goal. In contrast, recent work highlights how this property can be viewed as a strength: \citet{li2023eliciting} develop a methodology to use this mirroring to elicit, structure, and clarify users' thinking across various task domains. Recent work shows that efforts to mitigate sycophancy by considering pragmatic linguistic factors enable unifying sycophancy with other observed model failures \citep{cheng2026accommodation}.


 \paragraph{Deception} The emerging body of research on LLM deception increasingly focuses on measuring \textit{strategic deception}---defined as models ``deceiving selectively based on incentives or instructions'' \citep{jones2024lies}. While studies demonstrate that LLMs can produce deceptive statements in response to specific prompts, this work often faces two key interpretive challenges. First, it risks attributing observed behaviors to model \textit{intentions} to deceive. Second, results are often interpreted as evidence of model-level deceptive traits rather than instance- and context-specific behaviors. An alternative, less anthropomorphic framing, proposed by \citet{shanahan2023role}, views these behaviors through the lens of ``role-play'' where LLMs \textit{enact} human-like responses. This interpretation sees the system as context-bound, ``inferring and applying approximate communicative intentions'' \citep{andreas2022language}. Through this lens, complex behaviors like deception and self-awareness can be understood as sophisticated simulations rather than true cognitive states. This reframing also expands the set of interventions for deceptive behaviors: analyzing training data composition, examining how post-training interventions shape model behavior, and investigating reinforcement learning's effects on output distributions. 
 

\subsection{User interaction}\label{sec:user}
\textit{Assumption: Human-LLM interactions mirror human-human interactions.}
While this assumption can be helpful and reflects a common goal---for systems to be easy to use---its dominance can limit (1) users' ability to use LLMs effectively and (2) the types of LLM interfaces we choose to develop.

\paragraph{Human-like conversation as the dominant interaction paradigm}
The de facto interaction paradigm for human-LLM interaction is prompt-based interfaces, originally designed as debugging tools for machine learning engineers \citep{morris2024prompting}. As these interfaces resemble human-human chat interfaces, they may encourage users to naturally default to conversational patterns from human interaction. 
However, research on effective prompting suggests that optimal results often require structured, sometimes non-intuitive formats (e.g., ``least-to-most prompting'') rather than human-like communication patterns which rely on shared context and paralinguistic cues \citep{morris2024prompting, zhou2022least}. Simultaneously, research on human-LLM interaction shows that one of the key challenges users face is a significant \textit{gulf of envisioning} or ``distance between the human's initial intentions and their formulation of a prompt that foresees how LLM capabilities and training data can be leveraged to generate high-quality output'' \citep{subramonyam2024bridging}. This mismatch between natural dialogue and effective prompting requires greater experimentation with interaction paradigms and interface designs for LLMs. Structured interaction frameworks, using suggested inputs, guided flows, and/or domain-specific prompting strategies, would explicitly expose system capabilities rather than masking them behind conversational abstractions, effectively bridging the gulf of envisioning \citep{subramonyam2024bridging, feng2024cocoa, fagbohun2024empirical}. 

\section{Recommendations}
In the prior sections, we discussed concrete examples of non-anthropomorphic approaches in each step of LLM development. Here, we conclude with broader recommendations:
\begin{enumerate}
    \item \textbf{New metaphors for LLMs.} We encourage thinking beyond existing metaphors of LLMs toward new ones that capture the distinct qualities of LLMs. For example, \citet{shanahan2023role}'s ``role-play'' metaphor (and similarly ``agent models'' \citep{andreas2022language}), while employing folk psychological terms, carries conceptual precision that clarifies the unique characteristics of LLMs. In human-computer interaction (HCI), recent work has also illustrated possibilities of new interface metaphors that are less anthropomorphic \citep{so2026beyond}. 
    
    \item \textbf{Insights from cognitive science and linguistics.} In the cognitive sciences, 
    \citet{mccoy2023embers}'s ``teleological approach'' is useful in illuminating fundamental differences in how humans and LLMs operate and ought to be evaluated. Drawing on linguistics can also be beneficial: we can study LLMs' behavior as language, without assigning additional human-like characteristics, e.g., pragmatic theories can inspire new approaches to improve model capabilities \citep{cheng2026accommodation}.
    
    \item \textbf{Shifts beyond terminology.} While terminology is easiest to analyze, it need not be the only point of analysis or intervention. We hope that this paper is generative in illuminating underlying anthropomorphic assumptions. We encourage researchers to consider how these assumptions shape their work and how thinking beyond these assumptions can
 open up new paths for methodological development and theoretical frameworks.
    
\end{enumerate}

\section{Limitations}
Throughout this position paper, we have acknowledged the alternative view that anthropomorphic thinking is (1) natural and pragmatic, as well as (2) helpful. After all, anthropomorphism in LLM research serves important technical and social purposes. On the technical side, it provides intuitive frameworks for understanding complex systems and offers pragmatic terminology for discussing model behavior. It could also be argued that human cognition provides a proven template for intelligence, making it a valuable guide for AI development that has already led to breakthroughs. 
And, since LLMs are trained on human-generated data and designed to interact with humans, at least some anthropomorphic framing may be inevitable and even desirable. On the social side, anthropomorphic framing might improve our ability to engage non-technical stakeholders, communicate ethical considerations, and support policy discussions. Thus, we do not advocate for eliminating anthropomorphism entirely. Rather, we argue that awareness of the prevalence and limitations of anthropomorphic thinking can reveal new and potentially clarifying research directions.

\bibliography{custom}

\appendix

\section{Verb Analysis}
\label{sec:fightinwords}
To understand the cause of the prevalence of anthropomorphism, we perform a Fightin' Words analysis \citep{c} on our dataset (following the approach of \citet{cheng-etal-2024-anthroscore}), identifying verbs with highest z-scores for high ($>1$) and low ($<-1$) AnthroScore sentences, i.e., the words that differ most significantly in distribution between these two sets using weighted log odds ratios. This also enables us to understand the sensitivities and biases of the AnthroScore metric. The results are as follows:
\begin{itemize}
    \item  High AnthroScore: \textit{achieve, guide, demonstrate, teach, ask, train, prompt, follow, make, target, become, learn, understand, excel, require, mislead, answer, hallucinate, memorize, draw;}
    \item   Low AnthroScore: \textit{propose, outperform, use, develop, present, evaluate, enhance, improve, introduce, implement, validate, apply, reduce, adapt, employ, extend, allow, leverage, utilize, design}.

\end{itemize}
Top verbs for high AnthroScore reflect both human-like actions and also verbs that are commonly used in discussing LLMs, underscoring the prevalence of anthropomorphic terminology in papers about LLMs. 
\end{document}